# Enhancing streamflow forecast and extracting insights using long-short term memory networks with data integration at continental scales

Dapeng Feng, Kuai Fang[1], and Chaopeng Shen[2]

Civil and Environmental Engineering
Pennsylvania State University

**Abstract**

Recent observations with varied schedules and types (moving average, snapshot, or regularly spaced) can help to improve streamflow forecasts, but it is challenging to integrate them effectively. Based on a long short-term memory (LSTM) streamflow model, we tested multiple versions of a flexible procedure we call data integration (DI) to leverage recent discharge measurements to improve forecasts. DI accepts lagged inputs either directly or through a convolutional neural network (CNN) unit. DI ubiquitously elevated streamflow forecast performance to unseen levels, reaching a record continental-scale median Nash-Sutcliffe Efficiency coefficient value of 0.86. Integrating moving-average discharge, discharge from the last few days, or even average discharge from the previous calendar month could all improve daily forecasts. Directly using lagged observations as inputs was comparable in performance to using the CNN unit. Importantly, we obtained valuable insights regarding hydrologic processes impacting LSTM and DI performance. Before applying DI, the base LSTM model worked well in mountainous or snow-dominated regions, but less well in regions with low discharge volumes (due to either low precipitation or high precipitation-energy synchronicity) and large inter-annual storage variability. DI was most beneficial in regions with high flow autocorrelation: it greatly reduced baseflow bias in groundwater-dominated western basins and also improved peak prediction for basins with dynamical surface water storage, such as the Prairie Potholes or Great Lakes regions. However, even DI cannot elevate high-aridity basins with one-day flash peaks. Despite this limitation, there is much promise for a deep-learning-based forecast paradigm due to its performance, automation, efficiency, and flexibility.

Key points

1. We propose flexible data integration (DI), which can use various types of observations to improve discharge forecast at continental scales.

2. Adding a convolutional neural network unit to LSTM reduces overfitting but cannot outperform directly accepting lagged discharge as inputs.

3. Benefits of DI are strong in regions with high autocorrelation in streamflow, due to either groundwater dominance or surface water storage.

# 1. Introduction

Flooding is the biggest weather-related killer in the United States (NWS, 2014), while droughts incur on average $6B of damage every year in the US (NOAA, 2016). As the climate is predicted to bring more frequent extreme events for many parts of the US (Hirsch & Archfield, 2015; Stocker et al., 2013), accurate hydrologic predictions are of not only scientific value, but also great societal significance. In the US, hydrologic models such as the conceptual Sacramento Soil Moisture Accounting (SAC-SMA) model (M. G. Anderson & McDonnell, 2005; Burnash, 1973, 1995), among others (Krajewski et al., 2017; Maidment, 2017), have played major roles in supporting operational streamflow forecasting. These models have been extensively calibrated and tested over the US.

In the realm of process-based hydrologic modeling, data assimilation (DA) is a common method of utilizing observations to improve short-term streamflow forecasts at large scales (Clark et al., 2008; Houser et al., 1998; Vrugt et al., 2006). The main objectives of DA are to use recent observations to update model internal states, to better forecast future variables including observed and unobserved variables, and, less frequently, to update model structures or parameter sets. DA works by establishing the covariance between the internal states of a process-based model and observed variables, and using the difference between the observation and the model-simulated variable(s) to update the model internal states. Some variants of the DA algorithm can also help to correct model structure equations given some prior information (Bulygina et al., 2012; Nearing & Gupta, 2015). The uncertainty of the observation is considered through the covariance matrix. With the injection of new data, DA can remove autocorrelated effects of inevitable forcing errors and steer the model from incorrect trajectories caused by inadequate model structure or parameters.

Recently, time-series deep learning (DL) has emerged as a powerful and versatile modeling tool in hydrology (Fang et al., 2017; Kratzert et al., 2018; Shen, 2018; Shen et al., 2018), but few studies have examined streamflow forecast with DL, especially at large scales. DL models directly learn response patterns from massive amounts of data, without requiring manually-designed features or making strong structural assumptions (LeCun et al., 2015; Schmidhuber, 2015), and are hence less influenced by model structural errors. Our earlier work showed that the long short-term memory (LSTM) network (Hochreiter & Schmidhuber, 1997) can effectively learn soil moisture dynamics from satellite-based soil moisture products (Fang et al., 2017) and reproduce multi-year trends for root-zone soil moisture (Fang et al., 2018), with superior performance compared to simpler statistical methods. Since then, LSTM has already been utilized in a number of prediction tasks, including lake water temperature (Jia et al., 2019) and water table depth (Zhang et al., 2018). In particular, a number of studies have predicted streamflow using LSTM (Hu et al., 2018; Kratzert et al., 2018; Kratzert, Klotz, et al., 2019; Le et al., 2019; Sudriani et al., 2019). Most of these applications have focused on a few basins for their respective case studies, though Kratzert et al. (2018, 2019) utilized Catchment Attributes and Meteorology for Large-Sample Studies (CAMELS), a dataset with more than six hundred relatively undisturbed basins (Addor et al., 2017; Newman et al., 2014). While progress has been made, these studies appear to only scratch the surface of what could be achieved by time-series DL. In particular, there is significant potential for leveraging DL to flexibly absorb recent observations, whose value has not been fully exploited.

LSTM has often used lagged observations as inputs. For example, in the task of sentence completion, a partial sentence is provided as input to predict the next word, and when the new word is known, it is appended to the input sentence to predict the next word. Similarly, to predict streamflow at one gage, Hu et al. (2018) and Le et al. (2018) inserted streamflow observations an hour or a day before the prediction to improve LSTM forecast, which showed promising results. However, both of these streamflow studies trained the model on one gage without descriptors for the basin. While useful for the training basin, these models are not transferable to regions outside of the training one, and thus cannot learn from large-scale datasets. Consequently, a locally-trained model cannot help to derive hydrologic insights that depend on comparing and learning from regional patterns and their gradients. Because landscape parameters alter streamflow responses, it is an open question whether a uniform model could be trained to high performance for integrating recent observations at the CONUS scale.

If the observations are the most recent streamflow records, e.g., from yesterday, the forecast task is effectively to predict the daily streamflow changes due to various hydrologic processes that occur during the day, e.g., recession, new runoff, or baseflow return from early-season recharge. This daily prediction problem is arguably much simpler compared to the prediction without DI, which requires long-term memory and could be influenced by accumulated errors. Given this problem scope, it is not clear how advantageous LSTM would be compared to simpler methods such as an autoregressive model with exogenous terms, which has been successful at modeling recession and daily runoff processes (Vogel & Kroll, 1996).

Many sources of data, with various observational schedules (different revisit times, a single snapshot at a time vs. a multi-day average), are relevant to improving streamflow forecast. For example, some hydrologic stations report discharge only on weekly or monthly time intervals (Wang et al., 2009); the planned Surface and Water Ocean Topography mission will have recurrent snapshot measurements of around 11 days (Biancamaria et al., 2016). Operational constraints may often increase the latency between data acquisition and delivery. Furthermore, there are monthly-averaged satellite-based terrestrial water storage anomalies (TWSA) (Swenson, 2012) or soil moisture observations (snapshots), which are available on a 2-to-3-day revisit schedule (Entekhabi, 2010). The assimilation of data with various types, time scales, and latencies through traditional DA entails substantial expert supervision in selecting the most appropriate schemes for assimilation, data transformation (Clark et al., 2008), and bias correction. Especially since models often exhibit different behaviors from the observations, DA procedures often need bias correction to avoid distorting model internal states (Farmer et al., 2018). With hydrologic datasets, it has not been established whether LSTM can effectively utilize discharge observations with different types (multi-day average vs. single-day snapshot), latencies, and intervals (daily vs. weekly or monthly, etc.).

In this work, we tested an efficient and flexible LSTM-based method that automatically assimilates various types of discharge observations to improve the forecasts. We compared two versions of data integration (DI): one directly accepts lagged observations among the inputs, while the other passes a time-series segment of recent observations through a convolutional neural network (CNN) unit. DI refers to the general procedure of integrating recent observations to improve forecasts from a deep learning network. It is not a new concept, but has not been done at continental scales with large datasets, as discussed above. The use of the term "DI" allows for easier referencing and differentiation from DA, which is often used in forecasting tasks. DI does not utilize a pre-calibrated dynamical model as DA does. Rather, the method integrates the data injection and prediction steps into one step. Compared to the objectives of DA, DI alters the internal states of LSTM and improves the prediction of future predicted variables, but does not predict unobserved variables. For this work, we ask the following questions: (1) *Does LSTM with DI outperform reference statistical models for forecasting?* (2) *Can LSTM flexibly and effectively utilize snapshot, moving average, and regularly-spaced streamflow observations with different latencies for forecasting?* (3) *Where does LSTM perform poorly with or without DI, and what are the hydrologic processes behind such patterns*? In the following, we first describe the datasets, network, and DI method used, and demonstrate the effectiveness of DI compared to other hydrologic models, conventional statistical models, and LSTM without DI. We then show how LSTM was able to integrate different types of data. Finally, we interpret where and why LSTM with or without DI delivers strong or poor performance and relate the result to hydrologic processes. We show that besides improved estimates, big data machine learning could provide insights into how hydrologic systems function differently across the landscape.

**2. Data and Methods**

**2.1 Dataset**

We used the Catchment Attributes and Meteorology for Large-sample Studies (CAMELS) dataset (Addor et al., 2017; Newman et al., 2014), which consists of the basin-averaged hydrometeorological time series, catchment attributes, and streamflow observations from USGS for 671 catchments over the Continental United States (CONUS). Basins with minimal human disturbances were selected for inclusion in CAMELS. Most of the daily streamflow observations in CAMELS start from 1980 and end in 2014. The meteorological forcing data we used in this study is North America Land Data Assimilation System (NLDAS) daily data. The CAMELS catchment attributes that were used as inputs to our model include topography, climate characteristics, land cover, soil, and geology characteristics (Table 1). Unlike another LSTM-based study (Kratzert, Klotz, et al., 2019), we did not employ mean climate attributes as predictors, because we wished to test whether a competitive model could be trained using only physiographical attributes while using mean climate attributes as inputs would increase the risk of overfitting. Most of our results use all of the 671 catchments in CAMELS. Previous studies have noted either unclear watershed boundaries or too-large watershed areas with 140 basins, which are excluded from some evaluations (Newman et al., 2017). For the sake of comparison, we also report results for the subset of basins without these issues (531 basins).

Figure 1 plots eight attributes of the CAMELS basins including topography, basic climatic factors, baseflow index, 1-day-lag autocorrelation function of streamflow (ACF(1)), and the terrestrial water storage anomalies (TWSA) inter-annual variability index ($\gamma$). Baseflow index is the fraction of basin outflow attributed to baseflow based on a recession analysis (Ladson et al., 2013). $\xi$, the precipitation seasonality index, characterizes how much precipitation is seasonally in phase with energy inputs (Fang et al., 2016; Milly, 1994; Woods, 2009). It is positive when precipitation and energy are in phase, i.e., maximum rainfall occurs in the summer. $\xi$ is close to 0 when rainfall is evenly distributed, and is negative when precipitation and potential evapotranspiration are completely opposite in phase, i.e., most precipitation is in the form of winter snow. $\gamma$ is the ratio of variance explained by inter-annual variability and intra-annual variability for the TWSA data provided from the Gravity Recovery and Climate Experiment (GRACE) mission. Higher values of $\gamma$ indicate that the basin water storage has high inter-annual variability compared to intra-annual variability, i.e., variability between years is higher than variability within a year. Among the basin attributes, slope and soil depths are closely related as mountains tend to have shallower soils. $\xi$ is a climatic index while baseflow index and ACF(1) can be seen as hydrologic signatures.

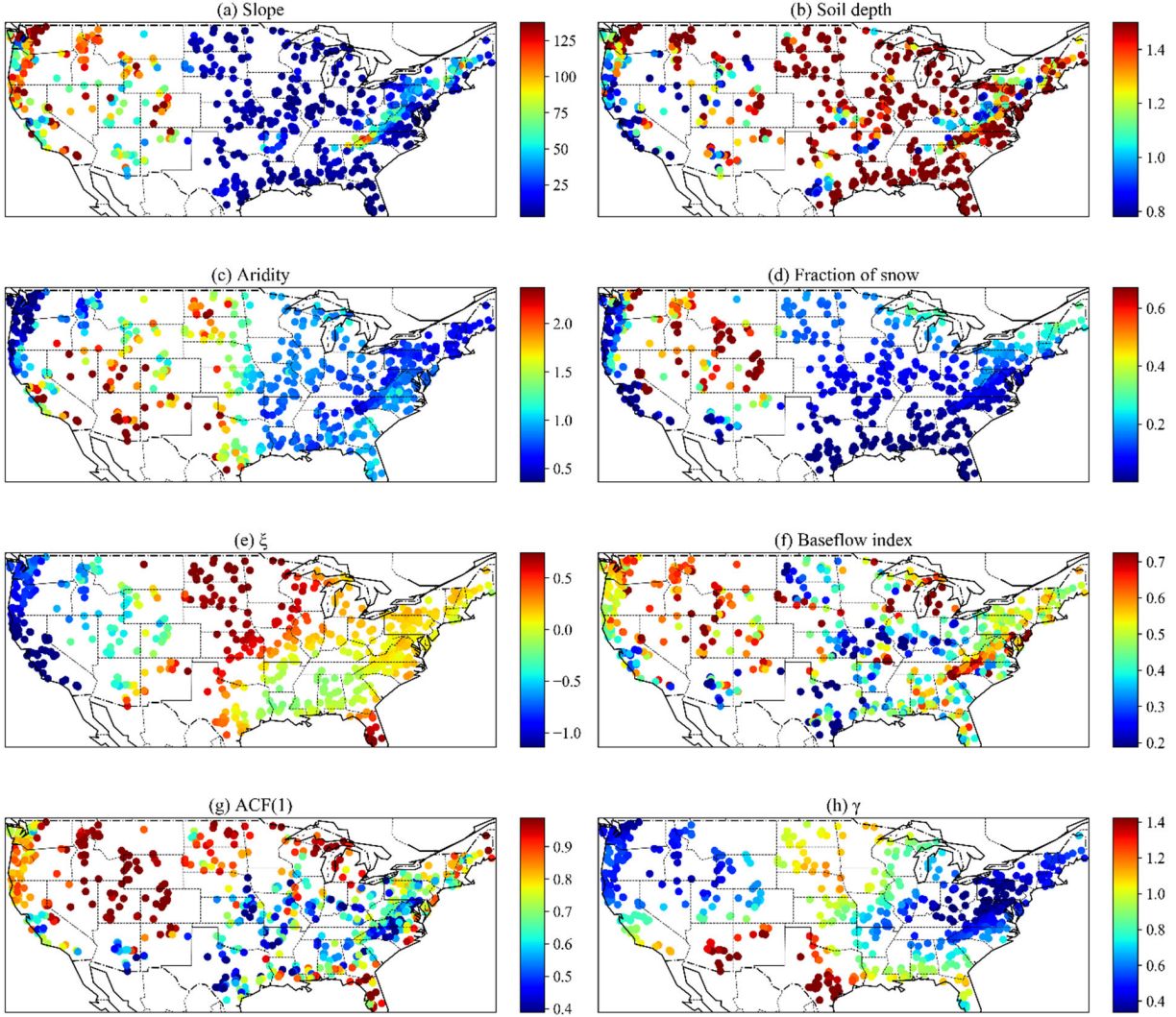

*Figure 1. The characteristics of CAMELS basins. (a) Slope: basin mean slope; (b) Soil depth (m); (c) Aridity: ratio between annual potential evapotranspiration and precipitation; (d) Fraction of snow: the fraction of precipitation falling as snow; (e) ξ, precipitation seasonality index: this index indicates the extent of which precipitation and energy inputs are in phase; (f) Baseflow index; (g) ACF(1): 1-day-lag autocorrelation function of streamflow; (h) γ, TWSA variability ratio: the ratio of TWSA inter-annual and intra-annual variability.*

## 2.2. LSTM model with Data Integration

We developed the LSTM streamflow model based on our previous soil moisture prediction work (Fang et al., 2017; 2018). LSTM is a type of recurrent neural network which learns from sequential data. Different from simple recurrent networks with one state variable, LSTM adds units such as "cell states" and "gates" (Hochreiter & Schmidhuber, 1997). The input, forget, and output gates control what information to allow in, what to forget, and what to output, respectively. These gates are all trained automatically and simultaneously, using input data to predict the target variable. This workflow is called "end-to-end" training, which avoids the need for expert-designed features. The memory cell enables LSTM to learn long-term dependencies such as snow and subsurface water storages, which are needed for streamflow predictions. Figure 2a illustrates the mechanisms within an LSTM cell. The forward pass of the LSTM model is described by the following equations:

| | | |
|---|---|---|
| Input transformation: | $x^t = ReLU(W_I I^t + b_I)$ | (1) |
| Input node: | $g^t = tanh(\mathcal{D}(W_{gx} x^t) + \mathcal{D}(W_{gh} h^{t-1}) + b_g)$ | (2) |
| Input gate: | $i^t = \sigma(\mathcal{D}(W_{ix} x^t) + \mathcal{D}(W_{ih} h^{t-1}) + b_i)$ | (3) |
| Forget gate: | $f^t = \sigma(\mathcal{D}(W_{fx} x^t) + \mathcal{D}(W_{fh} h^{t-1}) + b_f)$ | (4) |
| Output gate: | $o^t = \sigma(\mathcal{D}(W_{ox} x^t) + \mathcal{D}(W_{oh} h^{t-1}) + b_o)$ | (5) |
| Cell state: | $s^t = g^t \odot i^t + s^{t-1} \odot f^t$ | (6) |
| Hidden state: | $h^t = tanh(s^t) \odot o^t$ | (7) |
| Output: | $y^t = W_{hy} h^t + b_y$ | (8) |

where the superscript *t* represents the time step for time-dependent variables, $I^t$ represents the raw inputs to the model (Figure 2b), $x^t$ is the input vector to the LSTM cell, $\mathcal{D}$ is the dropout operator, *W's* and *b's* with different subscripts represent the gate-specific network weights and bias parameters, respectively, $\sigma$ is the sigmoidal function, $\odot$ is the element-wise multiplication operator, $g^t$ is the output of the input node, $i^t$, $f^t$, $o^t$ are the input, forget and output gates, respectively, $h^t$ represents the hidden states, $s^t$ represents the memory cell states, and $y^t$ represents the predicted output at time step *t*. We implemented a fast and flexible LSTM code that can utilize the highly-optimized NVIDIA CUDA® Deep Neural Network (cuDNN) library from the PyTorch deep learning platform. To reduce overfitting, $\mathcal{D}$ applies constant dropout masks (Boolean matrices to set some weights to 0) to the recurrent connections following Gal & Ghahramani (2015). We did not apply dropout to *g* in equation 6 as we did in previous work (Fang et al., 2018), as customization at this level in the code is not well supported by cuDNN.

A linear input transformation layer (equation 1) is used before the inputs are sent to the LSTM cell. This is our strategy to avoid important inputs being dropped out by the $\mathcal{D}$ operators, which can deteriorate the model performance. However, we recognize there may be other dropout formulations and this layer may not be necessary in those cases.

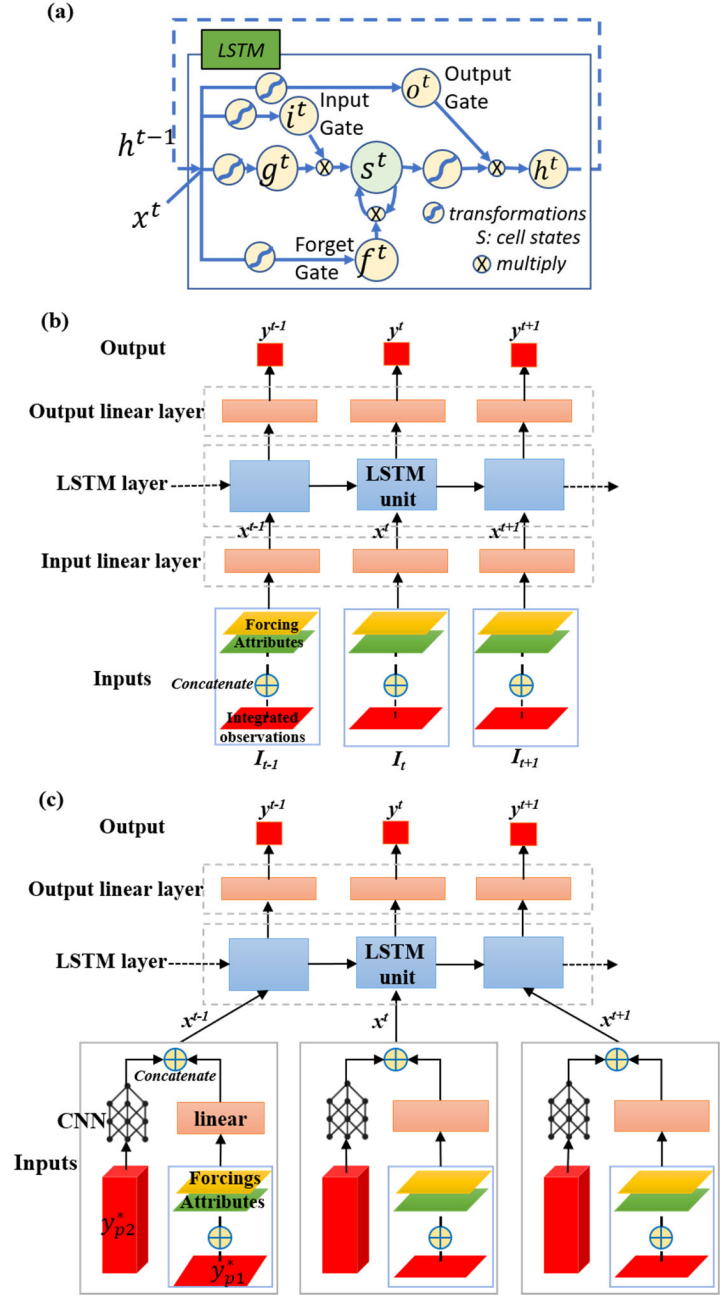

*Figure 2. The illustration of the LSTM unit and the data integration (DI) method. (a) The calculation inside an LSTM unit. The dashed arrow indicates that the hidden states from one time step are fed as inputs to the next time step; (b) The direct DI method without CNN unit; and (c) The DI method with CNN unit. Here the observation y\* is split into the directly used $y^*_{p1}$, and $y^*_{p2}$, the part passing through a CNN unit. The CNN unit is optional in that if the length of $y^*_{p2}$ is 0, e.g., in DI(1), this part of the network will not be used. For the projection model alone, neither $y^*_{p1}$ nor $y^*_{p2}$ would be used.*

Originally, the input of the base LSTM model included forcing and static attributes at the present time step. This base model without DI is referred to as the "projection model", as it does not use any historical observations. For the forecast models with DI, the base model inputs are supplemented with recent observations to improve output predictions for the current time step. These lagged observations can be directly supplied as inputs to the LSTM or passed through an operator that performs dimensional reduction, or both. To support DI, the inputs to the LSTM cell $x^t$ then become:

$$I^t = [x_o^t, y_{p1}^*] \qquad (9)$$

$$x^t = [ReLU(W_I I^t + b_I), R(y_{p2}^*)] \qquad (10)$$

where $x_o^t$ represents the original LSTM inputs including forcing and static attributes, $y_{p1}^*$ is the part of recently available observations that is directly assimilated as inputs, $y_{p2}^*$ is the optional part of the observation that will go through further transformation represented by the dimensional-reduction operator $R$, $x^t$ stands for the final inputs into the LSTM unit for equations 2-5, and $ReLU$ is the Rectified Linear Unit. The meanings of $y_{p1}^*$ and $y_{p2}^*$ are further explained in the following introduction to two kinds of integration experiments.

The network was trained on sequences of 365 days of 6 meteorological features and 17 static features (detailed in Table *1*) to predict the discharge at each time step. This fashion of training is sometimes referred to as "sequence to sequence." For the forecast model, discharge observations are integrated into the model at every time step following equations 9 & 10. At time step *t*, only observed discharges before *t* were used with DI.

(1) Direct integration

For most integration experiments in this research, the optional operator $R$ was not used and DI amounted to directly using different forms of historical observations as inputs (Figure 2b), such that equation 10 was the same as equation 1. The historical observations were directly added to the original LSTM inputs $x_o^t$ as the raw inputs $I^t$ shown by Equation 9. The directly integrated observation $y_{p1}^*$ can be of different types in our method, because given different types of observations, the network will adaptively discover the mathematical relationships between these observations and the prediction task. To test the hypothetical scenarios where different types of data are available, we tested integrating the following observation types: (1) a single daily observation that was acquired N days before the forecast, denoted by DI(N); (2) the moving averages of the previous N days' observations, DI(N)-M; (3) the regularly spaced snapshot observations with one observation in an N-day cycle, noted as DI(N)-R$^s$ (for DI(N)-R$^s$, one daily observation is integrated at a 1-day lag and there are no new observations for the next N-1 days, which is similar to real-world gages that have one snapshot reading every N days); (4) the average of observations in a regularly-spaced N-day cycle, noted as DI(N)-R$^a$; and (5) all the observations of the previous N days provided as separate inputs, noted as DI(N)-A. Table 2 provide a summary and some examples.

(2) Integrating observation time series through a CNN unit (operator $\boldsymbol{R}$ in equation 10)

If the integrated observations are long time series at each time step, the dimensional-reduction operator $\boldsymbol{R}$ can be used to extract useful features. We can flexibly choose the subsets of data to use as $\boldsymbol{y}^*_{p1}$ or $\boldsymbol{y}^*_{p2}$, which are collectively referred to as $\boldsymbol{y}^*$. $\boldsymbol{y}^*_{p1}$ contains the few most recent observations to be directly assimilated as inputs in equation 9, and $\boldsymbol{y}^*_{p2}$ contains the remaining part of the observation time series that will first be transformed by $\boldsymbol{R}$ before entering the LSTM cell (Figure 2c). For $\boldsymbol{R}$, we pass long historical records of streamflow as $\boldsymbol{y}^*_{p2}$ into a one-dimension (1D) CNN unit. The 1D CNN can accept a large number of inputs and reduce them to a small number of hidden layer outputs, which are subsequently fed into the LSTM cell, forming a CNN-LSTM network (Figure 2c). The rationale for introducing this unit is that CNN can extract important features such as recession rates or temporal gradients from the dataset and reduce the number of network weights, thus suppressing overfitting (see a small overview in Section 2.3 from the open-access article by Shen (2018)). If we attempt to integrate long records of past observations directly through LSTM, the number of weights in the input linear layer ($W_I$ in equation 10) increases with the number of inputs. This causes the fully-connected layer to become large, which is generally unfavorable as such layers are prone to overfitting. We hypothesized that adding the 1D CNN unit to the network would give the model more robust performance than directly sending all observations to LSTM, as tested with DI(N)-A. We refer to this CNN-LSTM network as CNN-DI($p_1$, $p_1$+$p_2$), a network that accepts recent $p_1$-day observations directly to LSTM's inputs and $p_2$ days of additional observations into the CNN units. For example, CNN-DI(1, 365) means that yesterday's observation ($y^*_{p1}$ in Figure 2c) is directly added to the original LSTM inputs at each time step, while the observations for the remaining 364 days ($y^*_{p2}$ in Figure 2c) are sent into the CNN unit and the extracted features are then fed into the LSTM cell. In this paper, $p_1$+$p_2$ ranged from 100 to 365 days, and the CNN coarsened the inputs to a hidden layer with a width of 3 to 10, respectively. The configuration of the CNN unit is described in Table S1 in the Supporting Information.

The input term $y^*$ ideally should have a similar physical meaning across time steps so the network can robustly learn its connection to the forecast target. Missing values occasionally exist in $y^*$ but present neural networks cannot handle NaN values. We also cannot interpolate to replace these values, because this would leak information from the future. Several potential treatments exist for handling missing $y^*$ values. First, a *closed-loop* version of the forward function in the neural network could be implemented to make predictions for the next time step and use them as inputs to the network, replacing those predictions with observations only when observations are available. This method could handle highly irregular and frequent missing values, but is computationally less efficient because it cannot utilize highly optimized computational libraries like cuDNN. Second, the network could be initially trained by using zeros to fill in for the missing values, and the trained network could then be used to make forecasts to fill the gaps. In theory, this procedure could be iterated a few times until the network converges. In practice, the streamflow data have very few isolated missing points (there are indeed whole periods with no discharge data, which are simply avoided by the loss function). We found the treatment scheme had little impact on the model predictions for this dataset. Therefore, we used the approach of filling zeros for the model training, which was the easiest and fastest option. For the forecast in the test period, a similar iterative method could be used to deal with missing data.

*Table 1. Summary of the forcing and attribute variables that were used as inputs to the LSTM model*

|  | **Variable Name** | **Description** | **Unit** |
|---|---|---|---|
| **Forcings** | PRCP | Precipitation | mm/day |
|  | SRAD | Solar radiation | W/m2 |
|  | Tmax | Maximum temperature | °C |
|  | Tmin | Minimum temperature | °C |
|  | Vp | Vapor pressure | Pa |
|  | Dayl | Day length | s |
| **Attributes** | elev_mean | Catchment mean elevation | m |
|  | slope_mean | Catchment mean slope | m/km |
|  | area_gages2 | Catchment area (GAGESII estimate) | km² |
|  | frac_forest | Forest fraction | - |
|  | lai_max | Maximum monthly mean of the leaf area index | - |
|  | lai_diff | Difference between the maximum and minimum monthly mean of the leaf area index | - |
|  | dom_land_cover_frac | Fraction of the catchment area associated with the dominant land cover | - |
|  | dom_land_cover | Dominant land cover type | - |
|  | root_depth_50 | Root depth at 50th percentile, extracted from a root depth distribution based on the International Geosphere-Biosphere Programme (IGBP) land cover | m |
|  | soil_depth_statgso | Soil depth | m |
|  | soil_porosity | Volumetric soil porosity | - |
|  | soil_conductivity | Saturated hydraulic conductivity | cm/hr |
|  | max_water_content | Maximum water content | m |
|  | geol_class_1st | Most common geologic class in the catchment basin | - |
|  | geol_class_2nd | Second most common geologic class in the catchment basin | - |
|  | geol_porosity | Subsurface porosity | - |
|  | geol_permeability | Subsurface permeability | m² |

*Table 2. Summary of all the data integration (DI) experiments in this study. Except for CNN-DI, which uses CNN-LSTM, all other models use LSTM only.*

| Experiment | Integrated observations | Example (the discharge at *t* is to be predicted) |
|---|---|---|
| LSTM | baseline LSTM model without using any observation | - |
| DI(N) | the lagged observation acquired N days ago | DI(1): adds the observation from day *t-1* |
| DI(N)-$R^s$ | the regularly spaced snapshot observation with N days' time interval | DI(3)-$R^s$: adds a single-day observation, regularly acquired once in every 3-day cycle |
| DI(N)-$R^a$ | the average observation of the last N days' time interval | DI(3)-$R^a$: adds the average observation of regularly-spaced 3-day cycles |
| DI(N)-M | the moving average of the previous N days' observations | DI(3)-M: adds the moving average of the observations from *t-3, t-2*, and *t-1* |
| DI(N)-A | all the observations of the previous N days | DI(3)-A: adds all of the observations from the 3 days' (*t-3, t-2, t-1*) |
| CNN-DI($p_1$, N) | all the observations of the previous N days using CNN-LSTM model | CNN-DI(1, 365): adds 365 days' observations at each time step; the observation from *t-1* is directly added to the LSTM inputs, while the observations from *t-365* to *t-2* are first sent to the CNN unit. |

To provide more details for the implementation, let us assume the time window for each training instance is $\rho$ and there are $n_f$ forcing variables. Then, the size of the forcing input is $\rho*n_f$, the size of integrated observation term for DI(1) through DI(N) is $\rho*1$. In other words, in each time step, there is only one value to be integrated. Same applies to DI(N)-M and DI(N)-R. For DI(N)-A, the size of the integrated observation is $\rho*N$. For CNN-DI($p_1$,N), the size of input for the $p_1$ part is $\rho*p_1$, the size for the CNN part is $\rho*(N-p_1)$.

The proposed model was trained from October 1, 1985 to September 30, 1995 and tested from October 1, 1995 to September 30, 2005. The objective function to be minimized, also called the loss function, was set as the root mean squared error (RMSE) between the predicted streamflow and the USGS observations, and the Adadelta algorithm (Zeiler, 2012) was used as the optimization method. We manually tested hyperparameter combinations and used a mini-batch size of 100, a hidden-state size of 256, and a training-instance length of 365 days as indicated in Table S1 in the Supporting Information. This length was substantially longer than our previous soil moisture prediction case in Fang et al. (2017) where lengths of 30 or 60 days were used. The longer length here can represent catchment snow and subsurface storage processes that have longer-term memory compared to surface soil moisture. It needs to be noted, however, that this length does not limit the memory of LSTM to 365 days. Considering the stochastic nature of the training procedure, we employed an ensemble of six simulations with different random seeds for each DI experiment. The ensemble-averaged discharge was also evaluated.

### 2.3. Data pre-processing

When trained on the entire CONUS, basins from the whole dataset are batched together to calculate the loss function. This batching method typically assumes the model errors are identically distributed among basins in the same batch. If no data preprocessing or normalization were applied, the default loss function would pay more attention to wetter and larger basins compared to drier basins. Here we tried several normalization procedures and chose the following steps, which provided optimal performance. We first normalized the daily streamflow by the basin area and mean annual precipitation to get a dimensionless streamflow value as the target variable, which reduced the differences between large and small basins. We then transformed the distributions of daily streamflow and precipitation, since these two typically have Gamma distributions:

$$v^* = log_{10}(\sqrt{v} + 0.1) \qquad (11)$$

where $v$ and $v^*$ are the variable before and after transformation, respectively. We aimed to make the transformed distributions as close to Gaussian as possible. We added 0.1 inside the log to avoid taking the log of zero. Finally, we applied a standard transformation to all the inputs by subtracting the CONUS-scale mean value and then divided by the CONUS-scale standard deviation.

### 2.4. Reference methods

To put the performance of our proposed DI method into context, we tested reference methods, including an auto-regressive model with exogenous inputs (AR) and a simple feedforward artificial neural network (ANN), in addition to the SAC-SMA hydrologic model provided from the CAMELS dataset. The equation of the auto-regressive model is:

$$y_t = c + \epsilon_t + \sum_{i=1}^{p} \alpha_i y_{t-i} + \sum_{j=1}^{r} \beta_j x_{j,t} \qquad (12)$$

where $t$ is the time step, $y$ represents the streamflow observations, $x$ contains the forcing terms, $p$ is the total order of the auto-regression model, $r$ is the total number of forcing variables, and $\alpha_i$ and $\beta_j$ are the estimated coefficients for the lagged streamflow observations and forcing terms, respectively. $c$ is a constant here, and $\epsilon_t$ represents white noise. A different AR model was trained for each basin. Therefore, the influence of different static attribute terms for each basin was absorbed into the AR parameters. Here we trained the model with $p$ equal to one for each individual basin, and denoted the resulting model by $AR_B(1)$.

We also ran a post-processing model in which the outputs of SAC-SMA were added to the above AR model as an additional regressor. This is similar to the concept of correcting forecast as a post-processing step to ameliorate model errors, which uses a range of methods including quantile-mapping (Wood & Schaake, 2008), linear regression, and autoregressive modeling (Bennett et al., 2016; Li et al., 2015). This reference model, named $AR_B^S(1)$, represents such post-processing methods, and will be compared with the proposed LSTM data integration.

The artificial neural network used here, noted as ANN(1), was a two-hidden-layer feedforward network whose inputs included the streamflow observation from the previous day. Unlike the $AR_B$ model, we trained one ANN model over all the CAMELS basins. The inputs to the ANN model are identical to that to DI(1), including the streamflow observations at the last time step. A typical calculation in a layer of the ANN model is:

$$S_t^k = f(W_k S_t^{k-1} + b_k) \qquad (13)$$

where $t$ is the time step, $k$ indicates the k-th linear layer, $W$ represents the weights, $S$ is the output, $b$ is the constant, and $f$ is the nonlinear activation function, for which we used $ReLU$ here.

Another potential model that we could evaluate is the least absolute shrinkage and selection operator (lasso) (Santosa & Symes, 1986). However, comparisons in our previous work showed that ANN was a stronger model than lasso (Fang et al., 2017). Therefore, we did not compare lasso to this work. Simulation results for SAC-SMA were downloaded from the CAMELS dataset (Newman et al., 2015), and we evaluated its performance within the same test period as our LSTM model for comparison.

**2.5 Evaluation metrics**

The metrics used to evaluate the performance of the models include percent bias and the Nash-Sutcliffe model efficiency coefficient (NSE) (Nash & Sutcliffe, 1970), all of which were calculated for each basin. We also report the percent bias of the top 2% peak flow range (FHV) and the percent bias of the bottom 30% low flow range (FLV) (Yilmaz et al., 2008). Here, considering the existence of zero flows, we did not calculate FLV in the log space as in Yilmaz et al., (2008). FHV and FLV highlight the performance of the model for peak flows and baseflow, respectively. The Pearson's correlation (corr) and Kling-Gupta efficiency (KGE) (Gupta et al., 2009) were used as additional evaluation metrics. KGE is a nonlinear combination of correlation, flow variability measure, and bias. All metrics are reported for the test period.

**3. Results & Discussions**

**3.1. Performance of LSTM projection and forecast models**

Without any data integration, projection LSTM already provided predictions that were competitive with other models, though there were still a few notable issues. The median NSE of the ensemble mean of 6 projection LSTM models for the 671 basins was 0.73 (Figure ), already higher than the value (0.64) from the SAC-SMA model (Newman et al., 2015) that is commonly used in operational flood forecasting (Figure **3**b and Figure 4b). The benchmark SAC-SMA simulations also had a longer calibration period (1980-1995), and parameters that were calibrated basin-by-basin. The CAMELS dataset included 10 different simulations of SAC-SMA and we chose the 6 best calibrated simulations to calculate the ensemble metrics. For the 531 subset basins (Newman et al., 2017), the median NSE values of the ensemble mean were 0.65 and 0.74 for the SAC-SMA and projection LSTM models, respectively. This comparison, consistent with Kratzert et al.

(2019), highlights that a CONUS-scale LSTM model was able to learn hydrologic behaviors across widely different basins without strong prior structural assumptions. We notice that the LSTM model shows good performance in the Northwest US (along the Rockies), in Northern California, along the Appalachian ranges, and in the Atlantic states (Figure 5a) (see a map of US states in Figure S1 in the Supporting Information). However, the projection LSTM model also had some weaknesses. Despite being noticeably stronger than SAC-SMA, the low flow component still had significant percent bias (Figure 3c and Figure 4c). 20% of the basins had a positive low flow percent bias (FLV) of 50% or more, and 8% of the basins had a negative FLV of -50% or more. Some basins even had more than 100% positive FLV. Most basins for which the projection LSTM model performed relatively poorly were concentrated along the Great Plains (see a map of US physiographic province in Figure S1 in the supporting information), which extends from North Dakota to northern Texas (Figure 5a). The other regions with relatively poor performance were along the southwestern border of the US, northern New Mexico and southern Texas, and the plains surrounding Lake Michigan, which are further discussed in Section **3.3**.

The projection model still has room for improvement: the hyperparameters were manually tuned, and we have not carried out a systematic hyper-parameter search. The distribution transformation (equation 11) could also be further optimized. We also did not consider factors such as the heterogeneities of topography and land use. Kratzert, Klotz et al. (2019) reported a slightly higher value of 0.76 for the median NSE (from the 531-basin subset) of the 8 ensemble-averaged discharge (as compared to 0.74 for our LSTM projection model using 6 ensemble members); the remaining difference between our models could be that we did not use average climatic conditions as inputs. Our goal was not to compete for the highest NSE values for the projection LSTM, but rather to highlight the effectiveness of the DI and understand where and why such effects exist.

We observe ubiquitous and heterogeneous benefits from DI over the CONUS. The median NSE of the ensemble mean improved to 0.86 and 0.80 after integrating the one-day-lag and three-day-lag discharges, respectively, with a reduced NSE variability (Figure 3b and Figure 4b). For the 531-basin subset, these two values are still 0.86 and 0.80 (Figure 4b). 96% of the basins over the CONUS benefited from the data integration, and the variance of bias was also greatly reduced. To the best of our knowledge, a median NSE of 0.86 is the highest number reported for daily streamflow forecast for hundreds of basins spread across the US. DI improved both the bias and the seasonality (as judged by correlation in Figure 3e) and is also higher according to KGE, which considers the flow variability measure. From the map, we could observe that most stations experienced a modest boost of 0.1-0.2 in NSE, but there were some regions with stronger improvements (Figure 5b-c). DI strongly boosted NSE values in the Great Plains from <0.4 to 0.7~0.95, except in southern Texas. The largest improvements were found in the Northern Great Plains (Region A on Figure 5c, hereafter referred to as F5-A; other regions are coded similarly), Central and Southern Great Plains (extending from south of F5-A to F5-E), Great Lakes region (F5-C), and Florida (F5-G). DI also substantially elevated NSE from 0.5~0.7 to 0.9 for the mid-latitude western states (F5-D), improving upon the already high NSE values there.

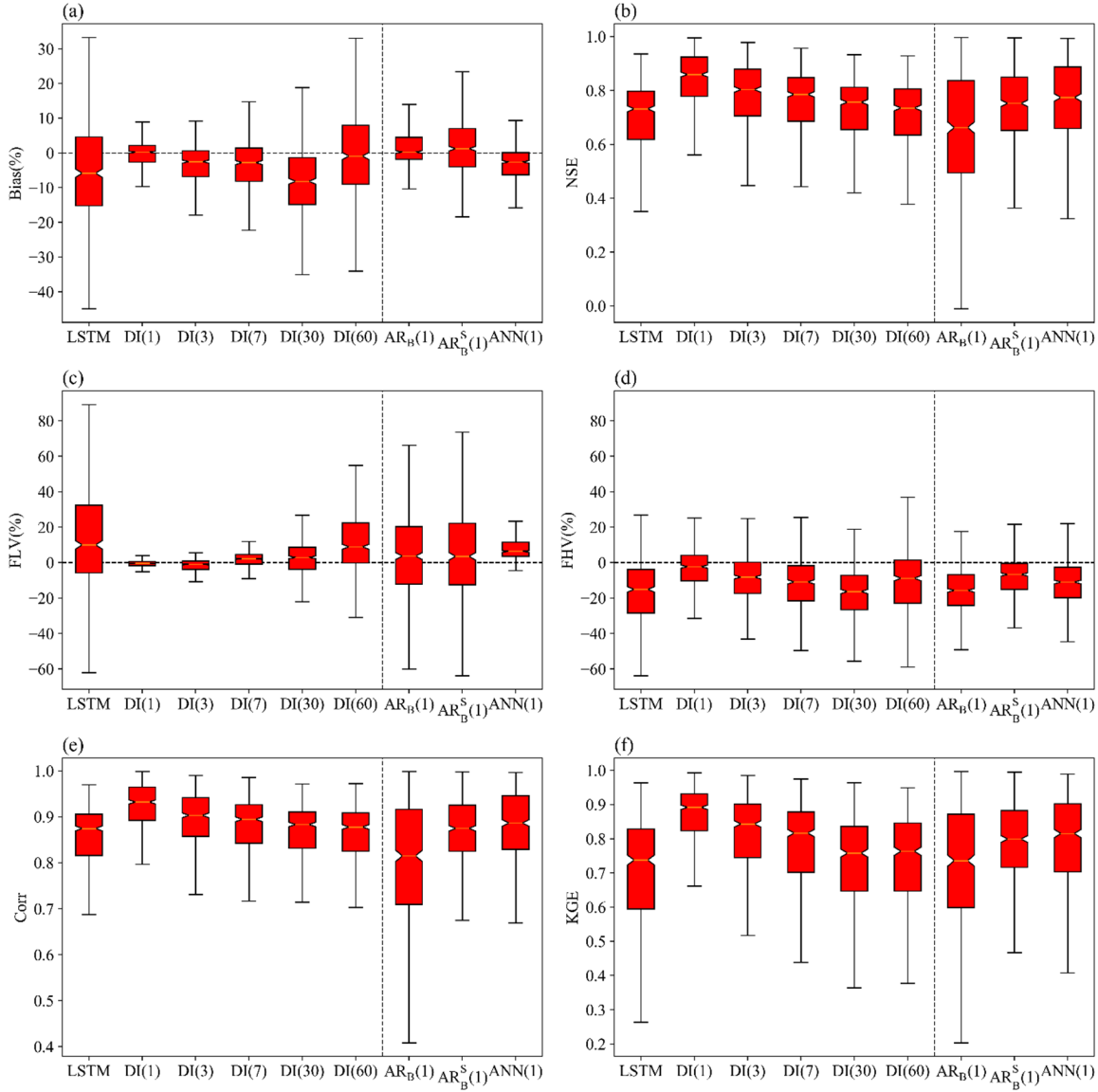

*Figure 3. Performance of the projection LSTM and DI(N) models in comparison with the $AR_B$ and ANN(1) models. DI(1) is an LSTM model that assimilates one-day lagged observations. All the metrics are calculated for 671 CAMELS basins based on the ensemble mean discharge with six ensemble members. (a) percent bias; (b) Nash-Sutcliffe model efficiency coefficient; (c) percent bias for the bottom 30% flow regime; (d) percent bias for the top 2% high flow regime; (e) Pearson's correlation; (f) Kling-Gupta efficiency.*

DI has larger benefits on a relative basis for the low-flow regime than for high-flow regime (Figure 3c-d, Figure 4c-d). The projection LSTM model could incur large bias for some stations, and for ~22% of the basins, the FLV exceeded ±50%. The FLV bias of the projection LSTM model was most likely due to the lack of subsurface characteristics (e.g., geologic layering, transmissivities, topography). The bias was mostly gone with DI(1). The reduction of overall bias, and especially FLV, is apparent for models from DI(1) to DI(7), and even to DI(30). The compaction of FHV due to DI, although still apparent, was not as large as that of FLV. This reduced effect is because peak flows have shorter time scales and are less dependent on memory compared to low flows. Nevertheless, the benefits of DI were still noticeable with FHV, representing the effects of antecedent conditions on flooding.

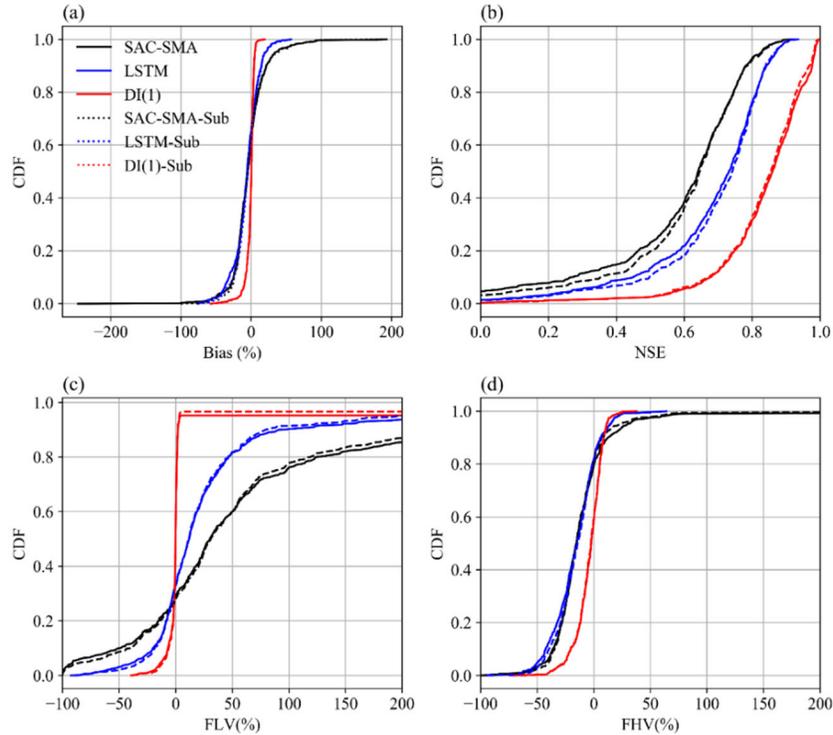

*Figure 4. Comparison of the cumulative density functions (CDF) for the DL models and the SAC-SMA hydrologic model. The "-Sub" suffix refers to the 531-basin subset used in previous studies, for comparison. FLV: The percent bias of low flow regime (bottom 30%); FHV: The percent bias of high flow regime (top 2%). The SAC-SMA model was calibrated from 1980-Oct-01 to 1995-Sep-30 and was evaluated from 1995-Oct-01 to 2005-Sep-30. LSTM and DI(I) were trained from 1985-Oct-01 to 1995-Sep-30, and the metrics were evaluated from 1995-Oct-01 to 2005-Sep-30. The CDF of FLV does not reach 1.0 because some basins have all zero flow observations for the 30% low flow interval, the percent bias can be infinite, and thus the x-axis cannot cover those basins. We set its x-axis to the same range as FHV.*

LSTM with DI exhibited substantial advantages over conventional statistical approaches including basin-specific $AR_B(1)$, $AR_B^S(1)$ and CONUS-scale ANN(1) (the three boxes to the right in all panels of Figure 3), all of which also used the previous day's discharge observations. $AR_B(1)$ and $AR_B^S(1)$ could forecast with a median NSE of 0.66 and 0.75, respectively (Table 3). By using outputs from SAC-SMA as inputs, $AR_B^S(1)$ was superior to the AR model, which showed the value of SAC-SMA and demonstrated that AR can be an effective post-processing strategy. Nevertheless, $AR_B^S(1)$ was still noticeably lower than DI(1), suggesting that the recurrent mechanisms in LSTM provided more functionality than autoregression. ANN(1) achieved the highest median NSE of 0.77 among the three reference models, but it still fell short of DI(1)'s NSE of 0.86. AR models had obvious bias according to FLV. Applying the transformation of equation 11 to AR models could mitigate the FLV bias, but it damaged the overall performance of basin-specific $AR_B$ models (i.e. the median NSE of $AR_B(1)$ reduced to 0.58 from 0.66 using this transformation), and was therefore not employed for $AR_B$ models here. $AR_B(1)$ tended to under-estimate the peaks and thus had negative bias for FHV. We suspect $AR_B(1)$ lacks the long-term memory that is needed to keep track of snow or subsurface water storages, which promotes sudden snowmelt peaks or saturated excess runoff. However, the addition of SAC-SMA simulations can enhance the ability to capture flow peaks, as shown by the much lower FLV bias of the $AR_B^S(1)$ model (Figure 3d). ANN(1) had a positive bias for FLV and a smaller negative bias for FHV compared with $AR_B(1)$ (Figure 3c-d), but when combining all flow regimes together, ANN(1) tended to have an overall negative bias (Figure 3a). The ANN was trained to minimize the overall sum of squared error. It balanced low flows and high flows, which produced positive FLV and negative FHV. The ANN had an even shorter memory than AR, and was likely overfitted to the peaks to keep the overall NSE down.

A CONUS-scale AR(1) model was also trained, but the results were worse than the basin-by-basin $AR_B(1)$ model (Table 3). Overall, AR lacked the complexity needed to learn from the large dataset and diverse responses observed over the CONUS. The $AR_B$ models with more regressors, e.g., $AR_B(365)$, also did not produce substantially different results, which suggests that the formulation of LSTM, rather than the number of memory storage units, is the primary difference between LSTM and AR. It was also notable that ensemble averaging greatly improved performance for the ANN(1), while the effects were more muted for the LSTM DI models.

The benefit of DI(N) decayed gradually as *N* increased, and the range of bias gradually widened (Figure 3a). Recall that this model used single-day observations from *N* days ago and for each *N* the model was trained separately. This gradual decay of DI benefits, to a certain extent, reflects the memory length of the hydrological processes. The value of information provided by observations about the near future wanes at larger time lags. When integrating the daily observation one month ago, i.e., for DI(30), the negative median bias of all the basins became worse than that of the projection LSTM model. However, the median NSE of DI(30) was still 0.76, higher than that of projection LSTM. For FLV, the benefits of DI are still clear at a 30-day lag (Figure 3c), presumably because recession flows have longer autocorrelation periods. On the other hand, FHV started to show a worse bias at 30 days. At a lag of about two months (60 days) or longer, the benefit of DI became less clear.

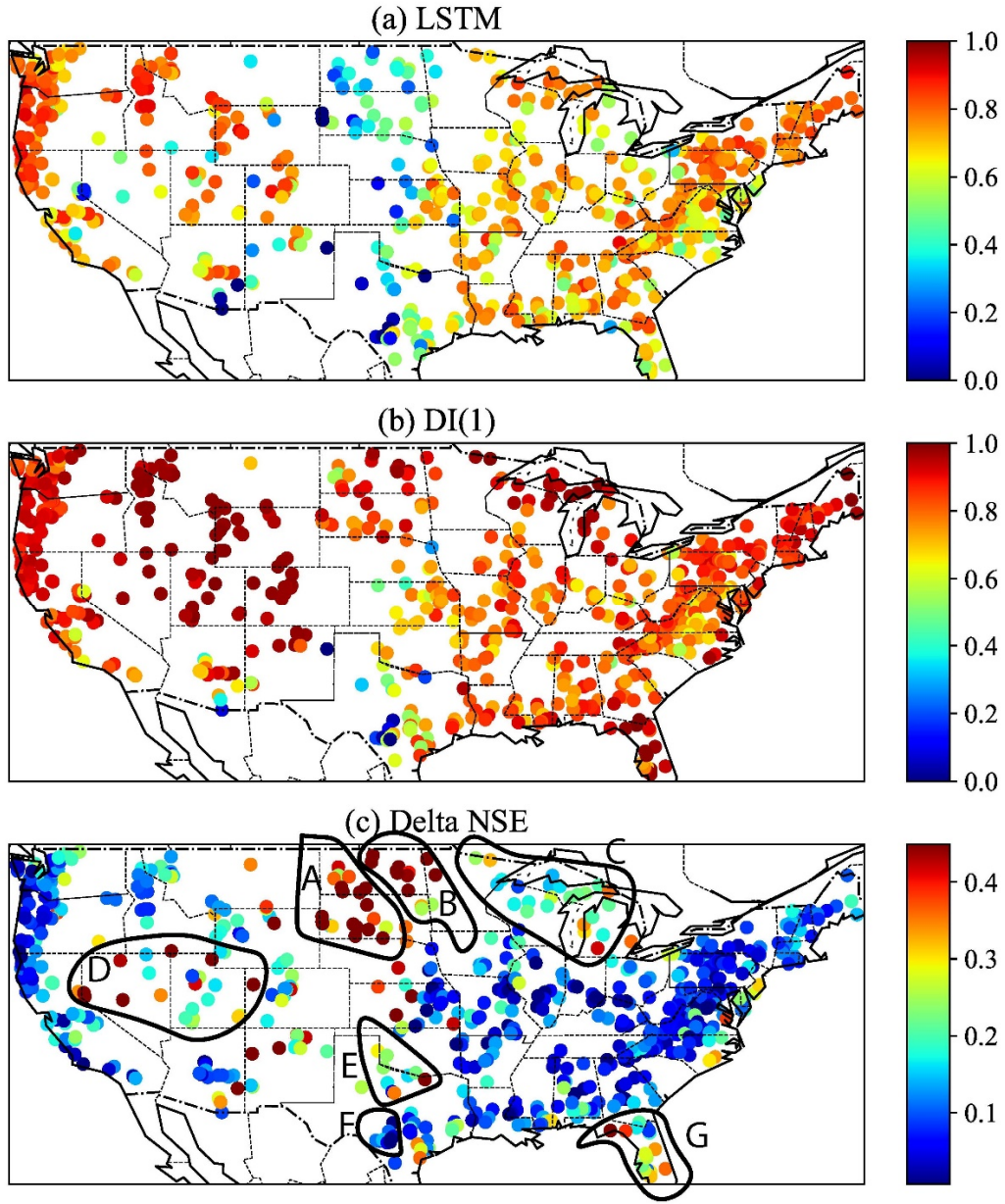

*Figure 5. Comparison of NSE spatial patterns over the CONUS (a) Projection LSTM without DI; (b) DI(1); (c) $\Delta NSE = NSE_{DI(1)} - NSE_{LSTM\_projection}$. Panel c has annotated regions for easy reference in the main text, though the boundaries are not precise for the related physiographies. In the text, we denote regions A, B, etc. on this panel as F5-A, F5-B, etc.*

In Figure 6, we show several examples of large performance differences between projection LSTM and DI(1), to better visualize typical issues with projection LSTM and how DI addressed them. Figure 6a and Figure 6b are two cases where DI significantly improved the bias with the baseflow, and their locations are marked in Figure 6l. For the basin shown in Figure 6a (in the Black Hills uplift, which is a mountainous area located inside the Great Plains in the southwest corner of F5-A), projection LSTM materially underestimated the baseflow, was not able to reproduce the flow peaks, and had a negative NSE. However, DI(1) fixed most of these problems: DI(1) provided strong prediction performance in both baseflow and peaks, and increased the NSE to 0.78.

For a basin in Michigan (Figure 6b), projection LSTM had a sustained positive bias for the baseflow. Previous research suggests that a major characteristic of the region is a thick glacial deposit layer, which induces groundwater-dominated streamflow (Niu et al., 2014; Shen et al., 2013, 2014, 2016; Shen & Phanikumar, 2010). It is possible that no model inputs distinguished these hydrogeologic characteristics from those of other regions, which otherwise would have allowed LSTM to differentially build a model for groundwater flow. DI(1) completely fixed the baseflow bias issue, and also returned a very high NSE of 0.96.

Although projection LSTM was able to adequately describe the baseflow for a basin in the Rocky Mountains (Figure 6c), it could not reach the observed magnitudes of peak flows. In two basins on the Great Plains and northern New Mexico, the projection LSTM model grossly overestimated the largest peaks (Figure 6d-e). For all these points, DI(1) more accurately reflected both the locations and magnitudes of peaks. Similarly, projection LSTM underestimated the peaks in a basin further north, in the Prairie Pothole Region (PPR, F5-B), where there was no baseflow (Figure 6f). For this basin, DI(1) helped to elevate the NSE from 0.43 to 0.92 by significantly improving performance at peaks. We will examine the hydrologic dynamics in the Northern Great Plains (NGP) and PPR and why we find large DI gains in these regions in Section **3.3**. Finally, for the point in South Texas, DI(1) did not improve the NSE. The hydrographs here show no baseflow and one-day flash peaks (Figure 6g, in F5-F). This region is discussed further in Section **3.3**.

The flow duration curves (FDC) for these basins are shown in Figure S2 in the supporting information. They also show that DI greatly improved the distributions of streamflow for most of the basins except for the Texas one. Except for c and e, the flow distributions simulated by LSTM were mostly closer to the observations than those from SAC-SMA.

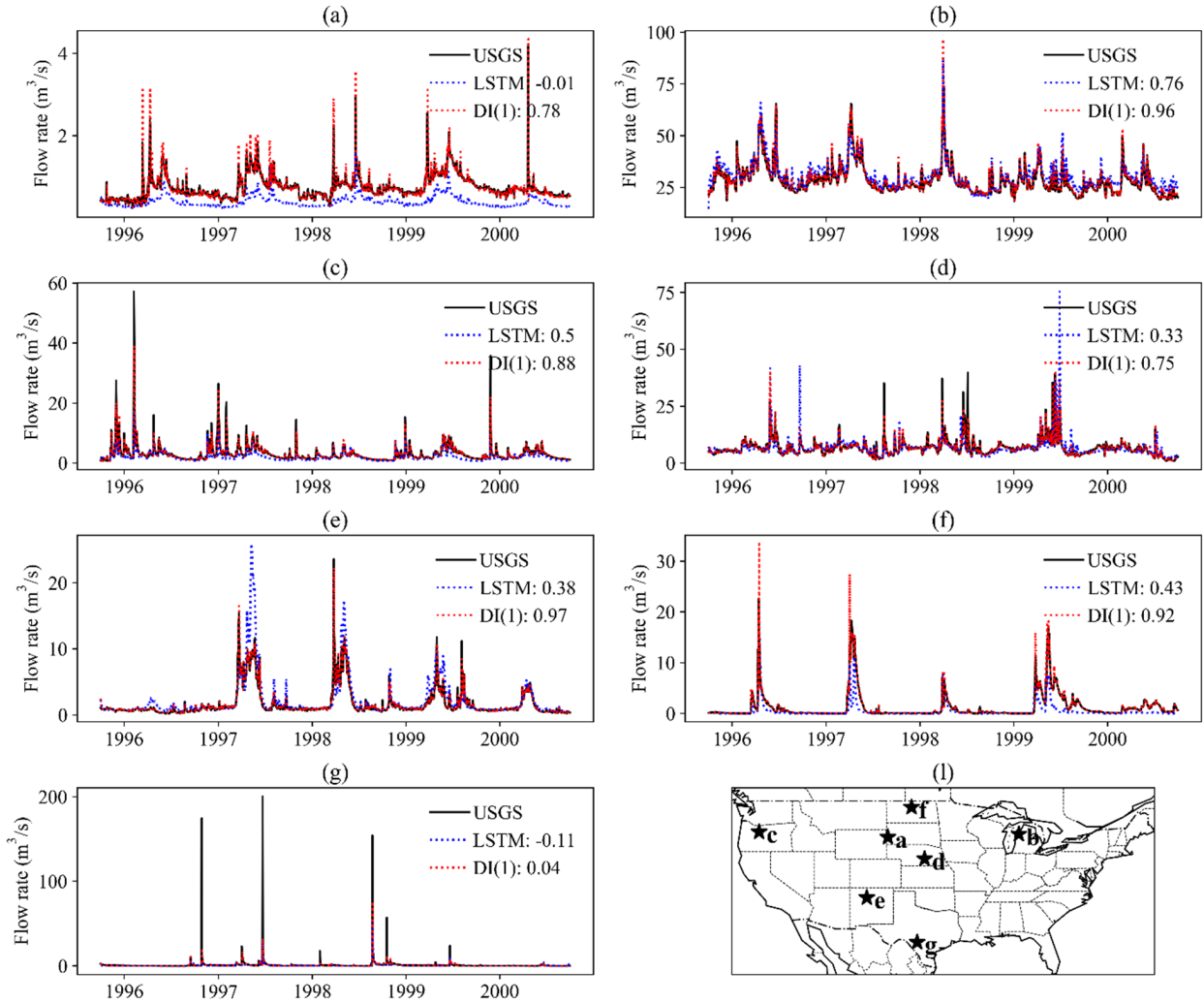

*Figure 6. Time series plots of selected basins to illustrate the benefits of DI to different flow regimes. The numbers in the legends represent NSE values of the simulation.*

**3.2 The flexibility of the proposed DL method for integrating different forms of observations**

Because of DL's special ability to automatically extract problem-relevant features and discover mathematical connections, it becomes procedurally trivial to assimilate various forms of data such as single-day, moving-average, regularly-spaced, and multi-day measurements. These different data sources proved to all be valuable, although their effectiveness varied (Figure 7). All the DI experiments in Figure 7 improved the forecast of LSTM models except when integrating the observations from one year ago, regardless of whether it was DI(N), DI(N)-$R^s$ or DI(N)-$R^a$. The NSEs were in the order of DI(N) < DI(N)-$R^a$ < DI(N)-$R^s$ < DI(N)-M < DI(N)-A. In other words, providing LSTM with observations from all of the previous days (DI(N)-A) was better than providing the moving averages from those previous days (DI(N)-M), which in turn was stronger than regularly-spaced snapshot measurements (DI(N)-$R^s$). These regular snapshots were generally more useful than averaged observations over the same period (DI(N)-$R^a$), while data from a single day with the same lag, DI(N), was the least useful.

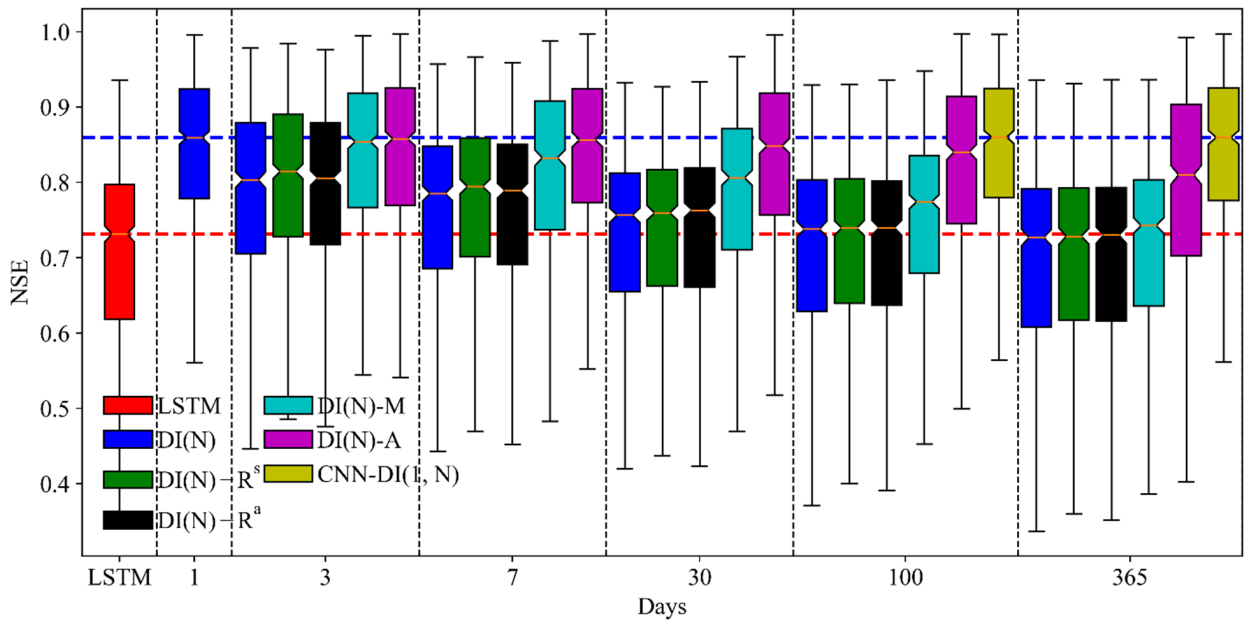

*Figure 7. Comparison of different data integration experiments as a function of lagged days. The red horizontal line highlights the median value of the projection LSTM model, while the blue horizontal line shows the median value of DI(1).*

Both DI(3)-A and DI(7)-A were quite close to DI(1). The gain of DI(3)-A compared to DI(1) was small (with a mean median NSE of 0.852 vs. 0.851, Table 3) and was not statistically significant when assessed using a Wilcoxon Rank Sum test. DI(1) was even slightly higher when we used the ensemble-mean discharge. CNN-DI(1,100) also only eked out a small and statistically non-significant lead (with a mean median NSE of 0.852) over DI(1). In contrast, DI(100)-A behaved noticeably worse than CNN-DI(1,100), even though they had the same information in their inputs; as expected, the DI(100)-A formulation was overfitted, while CNN-DI(1,100) demonstrated its robustness. Overall, considering the computational expenses required for CNN-DI, it can be argued that DI(1) is the better choice.

*Table 3. The CONUS-scale median NSE values of the ensemble experiments (6 ensemble members) for different data integration and comparison scenarios.*

| Experiment | Ensemble Statistics | | Ensemble mean discharge |
|---|---|---|---|
| | Mean | Standard deviation | |
| SAC-SMA | 0.636 | 0.0012 | 0.640 |
| LSTM | 0.714 | 0.0072 | 0.732 |
| LSTM-Sub | 0.722 | 0.0079 | 0.741 |
| DI(1) | 0.851 | 0.0097 | 0.859 |
| DI(3)-A | 0.852 | 0.0029 | 0.858 |
| DI(100)-A | 0.825 | 0.0088 | 0.840 |
| DI(365)-A | 0.785 | 0.0129 | 0.810 |
| CNN-DI(1,100) | 0.852 | 0.0037 | 0.860 |
| CNN-DI(3,100) | 0.852 | 0.0028 | 0.857 |
| CNN-DI(1,365) | 0.852 | 0.0039 | 0.860 |
| CNN-DI(3,365) | 0.850 | 0.0042 | 0.858 |
| AR(1) | 0.554 | - | - |
| $AR_B(1)$ | 0.662 | - | - |
| AR(365) | 0.613 | - | - |
| $AR_B(365)$ | 0.637 | - | - |
| $AR_B^S(1)$ | 0.752 | - | - |
| ANN(1) | 0.650 | 0.020 | 0.774 |

*"-Sub" refers to the performance on the 531-basin subset. "Mean" represents the average of the median NSE of 6 ensemble runs (median calculated across basins), and "Ensemble mean discharge" represents the median NSE of the averaged discharge from 6 runs. All the experiments were trained from 1985-Oct-01 to 1995-Sep-30 and evaluated from 1995-Oct-01 to 2005-Sep-30.*

There are two competing interpretations regarding the virtual equivalence of DI(1) and CNN-DI(1, 100): (i) The CNN unit did not extract any useful features from the long input series and was simply ignored by the LSTM; or (ii) the CNN unit extracted useful temporal gradient features, but these features were equally effectively extracted by LSTM via DI(1), so there was little room for improvement. Considering the detrimental impacts of DI(100)-A, it seems the CNN unit successfully carried out dimensional reduction and avoided overfitting. Therefore, we are leaning toward interpretation (ii). It is possible that LSTM learned to pass on injected observations to later time steps to construct useful features for forecasting. In summary, despite attempts to further improve performance, DI(1) is already a strong forecast model for this application and is difficult to surpass.

Regularly-spaced data with coarse temporal resolution is a common scenario, especially with satellite-based observations. Our results showed that this type of data, even at a sparse interval, offered the potential to improve forecasting accuracy compared to the projection LSTM model. LSTM is apparently capable of utilizing data with varied lags. For example, the integrated observation was only updated once every 30-day cycle for DI(30)-$R^s$ or $R^a$. However, because the latency of the assimilated observation changes throughout the cycle, the NSE of DI(N)-$R^a$ was lower than that of DI(N)-M. We attempted to add an additional input describing the number of days between the last new observation and the forecast time step, but it did not improve the forecast (data not shown here). Despite this, we still believe that the performance of DI(N)-R can be improved with future modifications.

There are several reasons that could explain the performance advantages of DI(N)-M over DI(N): (i) the moving average of the previous N days contained more recent information than one observation taken on the *N*-th past day; (ii) the moving average was a more appropriate measure to represent the history of the past days than a one-day measurement; and potentially (iii) the use of moving averages introduces less noise to the network. DI(N)-M also performs better than DI(N)-R, possibly because the data is more easily interpreted by LSTM. In reality, we seldom have moving average data without having the actual data from previous days, which, if available, would allow us to run the better-performing DI(1) instead. However, moving-average data would be preferred over DI(N)-R if a suitable temporal extrapolation scheme could be found, i.e., when regularly-spaced data could be re-gridded to the moving average periods. It is worth noting again here that these formulations were tested to show that these types of data could introduce benefits, but our choice will largely be constrained by what type of data is available.

There are at least two alternative hypotheses that could explain why integrating long-term-average discharge data are helpful compared to projection LSTM: (i) discharge carries information about forcing data in the distant past that was not otherwise available to projection LSTM; (ii) discharge better reflects the basin storage states. However, we ran additional DI experiments where we assimilated precipitation data rather than discharge data in DI(N)-M. The benefit over projection LSTM was minimal (data not shown due to its repetitive nature to the projection LSTM performance shown in Figure 7), meaning that hypothesis (i) mentioned above is not correct and hypothesis (ii) is more likely.

**3.3. Hydrologic insights through the lens of LSTM and DI**

In this section, we investigated conditions under which the projection LSTM model tended to work very well or poorly, and where DI helped substantially or did not. This knowledge gives us a glimpse into the hydrologic processes across the CONUS and the factors that control them.

**3.3**.1. The projection model performed well in mountainous and snow-dominated regions

Plotting single factors against NSE (Figure 8) showed that the NSE of projection LSTM is positively correlated with basin slope and fraction of snow, and negatively correlated with soil depth, aridity, precipitation seasonality ($\xi$), and inter-annual TWSA variability ($\gamma$). This suggests that LSTM is well suited to describe snow hydrology and streamflow in mountainous basins, which is consistent with the results shown by Kratzert et al. (2018). LSTM may do a good job at learning how to accumulate snow water in memory cells and release snowmelt water based on input forcing data (Kratzert, Herrnegger, et al., 2019). The projection LSTM model also did well in regions with negative precipitation seasonality (Figure 8e), which increases basin discharge given the same amount of precipitation. We also ran the analysis with the KGE, which showed similar patterns as the NSE (Figure S3 in the supporting information).

It is known that regions with stronger seasonality (which could be more negative $\xi$ in our case) are typically easier to model, leading to higher NSE for hydrologic models, conceptual or not. This is consistent with our observation with LSTM above, but we further found that even in these regions, LSTM can improve the NSE, although not as noticeably as other regions (Figure S5a in the Supporting Information).

**3.3**.2. DI was most beneficial in regions with high streamflow autocorrelation (ACF) caused either by high groundwater contributions or connected surface water storage

Figure 8o shows a strong positive correlation between NSE of DI(1) data points and streamflow ACF (Figure 8o). Basins with high ACF are where we can find maximal benefits of DI. This pattern could explain the strong ΔNSE for basins in the Northern Great Plains (NGP, F5-A) and scattered basins in Colorado, Nevada, and New Mexico (F5-D). Groundwater contributes to a large fraction of streamflow in F5-D, as evidenced by the high baseflow index (Figure 1g). In the Rocky Mountain region, such as western Montana, Idaho, Wyoming, and western Colorado, the projection LSTM model was already performing very well, with NSE mostly higher than 0.8, but DI still boosted the performance there, due to correspondingly large ACF values. In Nevada, Utah, and northern New Mexico, where ACF is still high but groundwater could have a very long travel time, we also found large benefits from DI.

Although high ACF often indicates large baseflow contributions and surface-groundwater connections, in some basins it does not. The Prairie Pothole Region (PPR, F5-B) contains basins with very little baseflow but high ACF, as shown in the example time series in Figure 6f. During rainfall hiatus in the PPR, water contributes to the potholes (or wetlands), which are disconnected from the streams, leading to a baseflow of nearly zero. After heavy storms, these wetlands could establish intermittent connectivity with each other and with the streams (Leibowitz et al., 2016; Leibowitz & Vining, 2003), primarily through surface routes (Brooks et al., 2018). These intermittent connections lead to dynamical contributing areas, making it challenging for projection LSTM to estimate flows. The wetlands serve as floodwater retention (Hubbard & Linder, 1986), introducing a higher flow autocorrelation on a daily time scale, which leads to a large improvement with the addition of DI. Besides the PPR, similar hydrologic dynamics could be found in the Upper Peninsula of Michigan (north of Lake Michigan) and Florida, although Florida has much more substantial groundwater contribution to streamflow. These are all regions with substantial water storage dynamics and thus strong autocorrelation in streamflow.

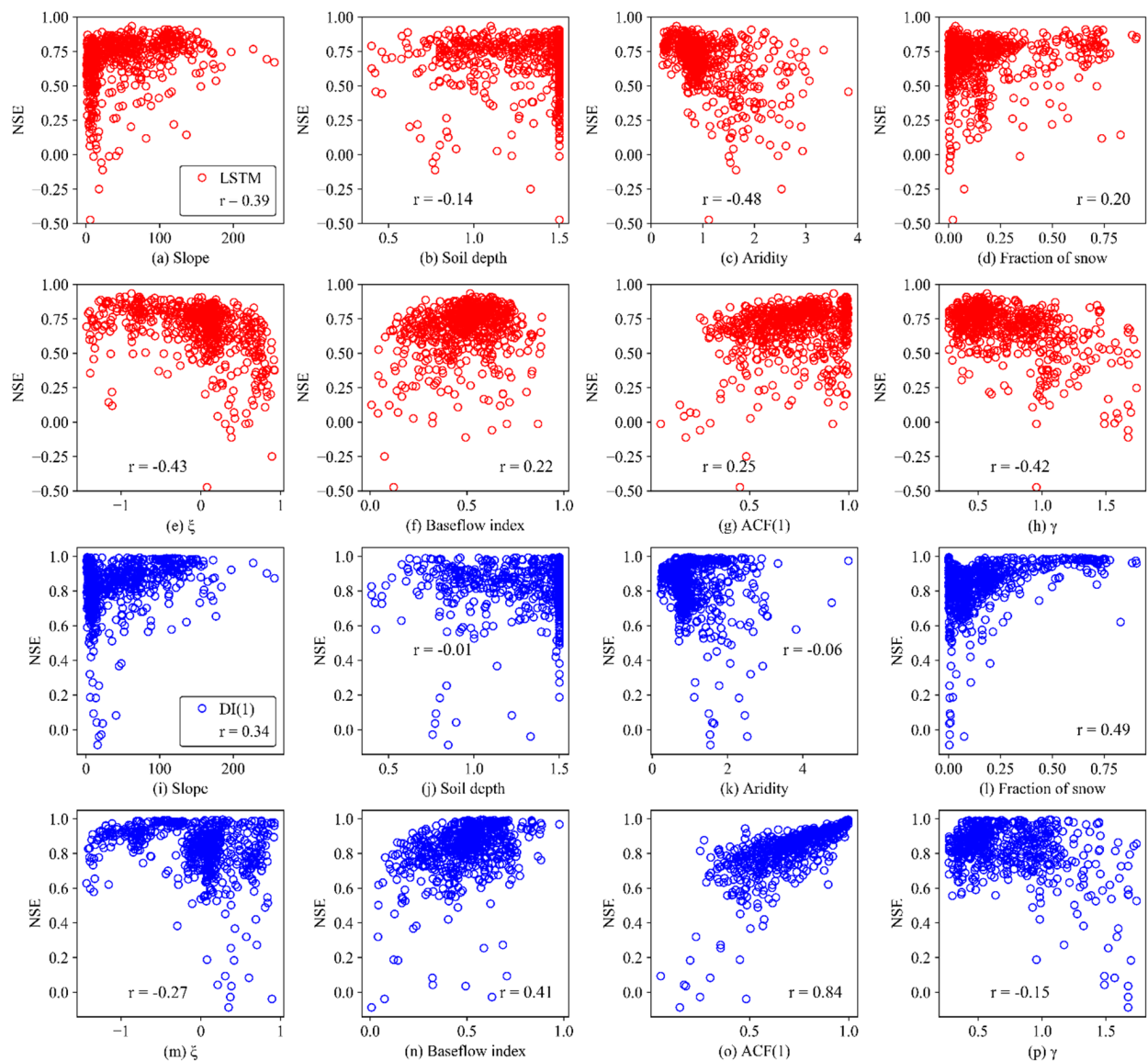

*Figure 8. Correlation between NSE and basin attributes over the CONUS. The panels with red points (upper two rows) are for the projection model, and the panels with blue points (lower two rows) are for DI(1). ξ: precipitation seasonality index which indicates how much precipitation and temperature are in phase; ACF(1): 1-day-lag autocorrelation function of streamflow; γ: the ratio of TWSA inter-annual and intra-annual variability.*

**3.3**.3. Even DI struggles with arid basins dominated by one-day flash floods

Near the US-Mexico border, i.e., southwest Texas (F5-F), southern California, and southern Arizona, both projection LSTM and DI(1) performed very poorly. These are highly arid basins with high inter-annual TWSA variability (γ), low ACF, sudden sharp hydrograph peaks (Figure 6g), and no storage effects. Adding to the difficulty of prediction, these sudden peaks may be strongly dependent on the minute-level rainfall intensity, which is not described in daily precipitation records, and could be a major reason why these peaks were poorly captured. The southwest Texas basins are also difficult to model because they are located on the karsted Edwards aquifer (characterized by porous bedrock with underground drainage systems), which promotes fast flows and inter-basin transfers. DI(1) likely did not help here because these flash peaks are one-day phenomena and have little relationship with yesterday's discharge.

**3.3**.4. Precipitation seasonality, aridity, and inter-basin transfers influence projection model performance but the errors could be corrected by DI.

NGP basins (F5-A) apparently caused trouble for the projection model, which is consistent with previous results from conceptual models that have historically reported poor performance in this region (E. A. Anderson, 2002). More recent evidence can be found in ABCD (Martinez & Gupta, 2010) and SAC-SMA (Newman et al., 2015) models. There are inherent challenges in predicting streamflow in this region, although no sufficient explanations have been offered. Poor performance is typically found from the NGP to the PPR (F5-B) in eastern North and South Dakotas. As the poor performance is spread over diverse landscapes, a general reason for this modeling difficulty may be due to the low annual basin discharge. For this semi-arid region, the precipitation seasonality that is synchronous with potential evapotranspiration (Figures 1e and 8e) increases hydrologic aridity (Berghuijs et al., 2014; Fang et al., 2016). As most precipitation arrives in the summer, when potential ET is at the maximum, it leaves little water available for runoff. For basins with small streamflow, a small bias in absolute magnitude would lead to a large drop in NSE. There could be a large number of dry days which could interfere with the training of LSTM. One might suspect that our normalization procedure by area and by precipitation (Section **2.3**) had sacrificed these large and dry basins when training a CONUS-scale model. However, we trained a model that was normalized by long-term average discharge, and the results were very similar (Figure S4 in the supporting information). Compounding the challenge, there is large inter-annual variability in TWSA on the Great Plains, and the initial storage states influence streamflow. It is difficult for projection LSTM to correctly initialize internal states related to water storages in the basin.

Nevertheless, there are also local reasons for the projection LSTM's difficulty in predicting streamflow for each landscape subtype. For example, in the Black Hills and immediate adjacent subregions (southwest basins inside F5-A), the hydrogeology is highly complex (Driscoll et al., 2002): orographic precipitation induces large spatial heterogeneity in rainfall and recharge in aquifer outcrops in the mountain peaks. Furthermore, spring flows, sinking streams, and incongruent surface and groundwater divides are common in this subregion, leading to inter-basin water transfers and frequent under-estimation of peaks. In this region, groundwater has a strong influence on the streamflow, leading to a high ACF and therefore large benefits from DI. Near the center of the South Dakota-Nebraska border (southeast edge of F5-A), the rivers could also receive large groundwater contributions from the Ogallala and Sand Hills aquifers (Andrew J. Long et al., 2003), resulting in large ACF values, and correspondingly large benefits from DI.

The scatter plots in Figure **9** clearly show that basins with low NSE are also mostly basins with low mean annual streamflow and high precipitation-energy synchronicity. The mid-latitude western states (F5-D) have similar annual discharge volumes as their southern neighbors (in the region with $\gamma < 0.75$ and annual runoff < 200 mm/yr in Figure **9**). However, they have lower inter-annual TWSA variability ($\gamma$), indicating more water passing through the subsurface system and more groundwater contribution to channels. With correspondingly higher ACF values, predictions for these regions are greatly enhanced by the addition of DI. We see the advantage of LSTM over SAC-SMA decreases for basins with large inter-annual variability (Figure S5b in the supporting information). Figure 9 suggests that we can anticipate general reliability of the LSTM and DI(1) models prior to modeling, using climatic factors and TWSA observations, although further validation is needed from other continents.

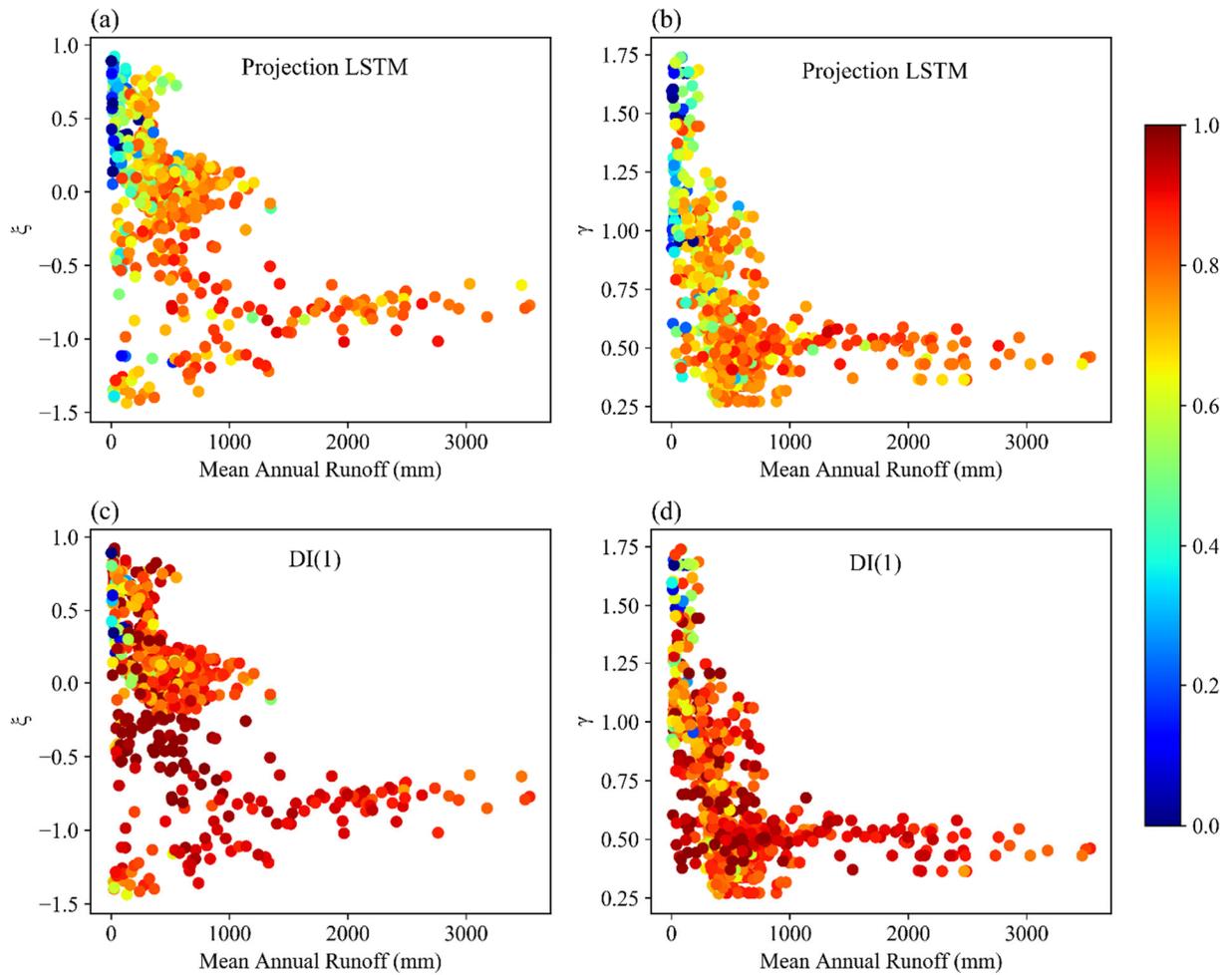

*Figure 9. Scatter plots of NSE as functions of long-term mean streamflow, precipitation seasonality, and inter-annual storage variability. ξ : precipitation seasonality index which indicates how much precipitation and temperature are in phase; γ: the ratio of TWSA inter-annual and intra-annual variability. The color in the upper and lower panels indicates NSE values for projection LSTM and DI(1), respectively.*

**3.3**.5. Summary

Overall, projection LSTM had difficulty with baseflow bias and regions with low runoff volumes, strong heterogeneity, and prevalent wetlands or lakes. The latter two cases are difficult for most conceptual hydrologic models to capture, as they differ from the assumptions of the standard rainfall-runoff processes that these models are based on. It will be difficult to improve the performance of projection LSTM in these basins unless we can provide more detailed knowledge in the inputs, such as hydrogeologic information. However, if the streamflow has a high ACF, it can be greatly improved, as it was in this study, by the integration of recent observations. High ACF could be due to either more prominent baseflow, such as the arid west or Black Hills region of the NGP, or the presence of surface water storage and intermittent connectivity as in the PPR, in which case the peaks could also be improved. The NSE of LSTM, with or without DI, could be low for very arid basins with one-day flashy peaks and large inter-annual TWSA variability, which have less water passing through the subsurface. Analyzing these examples helped to highlight the strengths and limitations of LSTM and DI. Additionally, the behavior of LSTM also shed insights into related hydrologic processes. For example, a large benefit from DI would suggest a strong influence from slow hydrologic processes such as large groundwater and surface water storages.

### 3.4. Further discussion and future work

There may be debates regarding whether this technique should be called data assimilation, data integration, autoregression, or other names. DI does not separately use a pre-trained forward model, and also does not predict unobserved quantities, which are features of DA. DI does match some of the primary objectives of DA, though, including the utilization of recent observations to update model internal states and improve forecast. Moreover, with the same setup, it should be possible to integrate other related variables, such as recent observations of soil moisture, GRACE water storage anomalies, or canopy states, to further improve streamflow forecast. Additionally, LSTM has the potential to evolve into a fully functional hydrologic model and DI may evolve into many variants. These kinds of functions are far more than what conventional autoregression is typically known to offer. This flexible scheme also opens up new possibilities, including assimilating multi-day observations at the same time. Thus, we think a separate term, data integration, is justified.

Integrating different forms of data was straightforward and uniform with LSTM. For this to be done with traditional data assimilation, different assimilation schemes, different covariance matrices, and varied bias correction schemes would have been required. For example, to assimilate monthly TWSA data from the GRACE mission, an ensemble Kalman smoothing filter algorithm needs to be applied (Reager et al., 2015; Zaitchik et al., 2008). For observations of different scales, multiscale schemes need to be employed (Pan et al., 2009). The different schemes required for different data sources and resolutions could add to the complexity of the forecast system. For the deep learning (DL) system, the scheme was almost uniform for all data forms, as DL handled the mathematical details. This simplicity allows for the liberation of human minds from mathematical details, to instead focus on questioning, problem formulation, and data collection.

The CNN-LSTM did not produce any benefits compared to the simple DI(1) model in this case. This is partially because DI(1) is already very strong; LSTM may already be able to pass observations in multiple steps to construct gradient-like features. However, we envision future scenarios where CNN-LSTM could have value, as it has been shown to reduce overfitting when there are substantial amounts of raw data that could not be directly passed into LSTM. For example, the influence of topography could be introduced this way.

The current scheme does not consider the uncertainty of the observations. For example, measurement uncertainty should be higher under extreme peak flow conditions. Typical DA or data fusion schemes will incorporate such information through an update formula. Recently, we have examined new methods to estimate LSTM model uncertainties which have estimated predictive error well, especially for temporal extrapolation (Fang et al., 2019). In the future, it should be possible to add this uncertainty estimation into the data integration scheme. Moreover, in that recent work, we also explored an option to have LSTM build an error model to estimate the error based on available inputs. Indeed, our analysis in Section **3.3** suggests that errors are dependent on many input attributes, and these relationships could be utilized by LSTM to build error models. Nevertheless, systematic errors with the observations themselves cannot be assessed by this method or any other data-driven method.

It can be said that the proposed DI scheme learns both the streamflow generation process and the procedure of using observations to improve forecast. We were able to obtain hydrologic insights by examining the performances of projection LSTM and DI. However, it remains challenging to understand what LSTM has extracted or computed. Continued effort in interpretive machine learning is encouraged to shed light to the mechanisms of the LSTM updates, e.g., regarding how multi-day data are employed by CNN-LSTM and how LSTM has derived its own snow storage and melt mechanisms, which could help us design better hydrologic models.

We see that the projection LSTM model, while it is already performing better than many operational flood prediction models at the CONUS scale, still encounters issues that seemingly could have been resolved without employing near-real-time DI. For example, for basins with significant baseflow or inter-annual storage variability, the issue with initial states could be addressed by providing one measurement at the beginning of the simulation, baseflow conditions such as monthly-average streamflow, or even expert and non-quantitative information. Such information that is difficult to utilize in a process-based model could be leveraged in DL flexibly, as long as it could be provided in sufficient quantities for training. As discussed earlier, we suspect that many of the baseflow biases are due to inadequate geologic descriptions for the subsurface, and would require additional input data.

While DL has shown great promise here, we do not advocate it as a silver bullet that solves all problems. This work clearly showed that the projection model performed poorly on the Great Plains, just as previous conceptual models have, and it clearly had limitations for the Prairie Potholes Region before the addition of DI. DL cannot create relevant information out of thin air, nor can it distinguish between causal and associative relationships. While our earlier work showed that LSTM was able to project long-term trends in soil moisture (Fang et al., 2018), DL's capability for long-term projection for streamflow needs to be carefully evaluated. We see abundant synergies between DL and process-based models (Shen et al., 2018). In the future, process-based models could be coupled to DL models by, for example, providing training priors, conditioning network weights, or constraining loss functions, as outlined by Karpatne et al. (2017), to enable more reliable future projections and predictions in ungauged basins. It is generally recognized that adding physical constraints could improve the robustness of machine-learning-based predictions, an argument we are in agreement with. Although not specifically implemented in this work, we think that coupling process-based model elements with DL will likely be on the list for future work.

## 4. Conclusions

In this paper, we used LSTM to integrate various types of recent streamflow observations to improve streamflow forecast performance. While the use of lagged streamflow has been demonstrated for single-basin training data in the past, no study previously showed the effects on forecasts at continental scales or with basin attributes. Consistent with literature results, projection LSTM without DI showed results competitive with extensively-calibrated operational models, especially in mountainous and snow-dominated basins. However, like other types of hydrologic models, projection LSTM had issues with baseflow bias, basins with large inter-annual storage changes, hydrologically arid basins (either due to small annual precipitation or due to semi-aridity coupled to in-phase precipitation seasonality), and regions with inter-basin transfers and complex hydrogeology.

After applying the DI procedure, the model was able to address most of the abovementioned issues, producing an unprecedentedly high national-scale forecast NSE value (0.86, from the ensemble mean discharge of DI(1)). DI provided wide-spread and spatially-varying benefits, and the largest gains were obtained for basins with strong flow autocorrelation, which suggest either strong storage-surface connections or surface water retention. Model performance was especially elevated in the Great Plains, northern Texas, the Great Lakes region, and in semi-arid mid-latitude western states. DI can improve both baseflow and peak predictions for regions where peaks are induced by varying surface-water connectivity such as the Prairie Pothole and Great Lakes regions. LSTM with DI was substantially stronger than simpler statistical models, e.g., auto-regressive models or simpler feedforward neural networks, for both peak flow and baseflow portions of streamflow. However, one region that even DI was not able to improve was in southern Texas, an arid region with karst hydrogeology and flashy streamflow peaks with no baseflow.

We found that this network can flexibly integrate lagged, multi-day, moving average, and regularly-spaced snapshot or time-averaged observations with a uniform structure. All of these sources of data allow the network to perform better than the base projection LSTM model. Such benefits were not simply because of the long memory of the forcings, as assimilating precipitation data did not provide any performance gain. The flexible LSTM model automatically learned how to approximate the mathematical operations for both the hydrologic process and the model-data integration procedure. This capability changes the way we ask questions.

The more complicated CNN-LSTM architecture could not deliver statistically better performance than simply integrating 1-day-lag observations as inputs to LSTM. This result is in general agreement with the literature, where modifications to LSTM structure have not led to performance gains. The lesson here is that, given the same sequential input information which could be utilized by LSTM, it may be difficult to design an architecture that surpasses LSTM with significant margins.

**Acknowledgments**

The CAMELS dataset, including catchment attributes, forcing data, and streamflow, can be downloaded from the citations provided in this paper. This work was supported by National Science Foundation Award EAR-1832294. KF was partially supported by Office of Biological and Environmental Research of the U.S. Department of Energy under contract DE-SC0016605. GRACE TWSA data can be downloaded from GRACE monthly mass grids (https://grace.jpl.nasa.gov/data/get-data/). Streamflow data can be downloaded from U.S. Geological Survey Water Data for the Nation website (http://dx.doi.org/10.5066/F7P55KJN). Our LSTM code is available at GitHub (https://github.com/mhpi/hydroDL). We appreciate phone conversations with Dr. Daniel Driscoll from the USGS regarding hydrologic processes in the Dakotas. Many thanks to Ms. Kathryn Lawson who helped to proofread the manuscript.